\documentclass{article}

\usepackage[preprint]{acl}

\usepackage[utf8]{inputenc} 
\usepackage[T1]{fontenc}    
\usepackage{hyperref}       
\usepackage{url}            
\usepackage{booktabs}       
\usepackage{amsfonts}       
\usepackage{nicefrac}       
\usepackage{microtype}      
\usepackage{xcolor}         
\usepackage{graphicx}
\usepackage{tikz}
\usepackage{amsmath}
\usepackage{tabularx}
\usepackage{makecell} 
\usepackage{multirow}
\usepackage{booktabs}
\usepackage[most]{tcolorbox}
\tcbuselibrary{listings,skins,breakable}
\usepackage[normalem]{ulem} 
\usepackage{float} 
\usepackage{amsmath}

\title{$\textbf{OThink-SRR1}$: Search, Refine and Reasoning with Reinforced Learning for Large Language Models}

\author{
 \textbf{Haijian Liang\textsuperscript{1}},
 \textbf{Zenghao Niu\textsuperscript{1}},
 \textbf{Junjie Wu\textsuperscript{2}},
 \\
 \textbf{Changwang Zhang\textsuperscript{2}},
 \textbf{Wangchunshu Zhou\textsuperscript{2}},
 \textbf{Jun Wang\textsuperscript{2}},
\\
 \textsuperscript{1}Shenzhen University, \\
 \textsuperscript{2}OPPO Research Institute
\\
changwangzhang@foxmail.com
}

\begin{document}
\maketitle

\begin{abstract}

Retrieval-Augmented Generation (RAG) expands the knowledge of Large Language Models (LLMs), yet current static retrieval methods struggle with complex, multi-hop problems. While recent dynamic retrieval strategies offer improvements, they face two key challenges: 1) irrelevant retrieved noise can misdirect the reasoning process, and 2) processing full documents incurs prohibitive computational and latency costs.
To address these issues, we propose OThink-SRR1, a framework that enhances large models with an iterative Search-Refine-Reason process trained via reinforcement learning. Its core Refine stage distills retrieved documents into concise, relevant facts before reasoning. We introduce GRPO-IR, an end-to-end reinforcement learning algorithm that rewards accurate evidence identification while penalizing excessive retrievals, thus training the model to be both focused and efficient.
Experiments on four multi-hop QA benchmarks show our approach achieves superior accuracy over strong baselines while using fewer retrieval steps and tokens. This positions OThink-SRR1 as a potent foundational model for information-seeking agents.

\end{abstract}

\section{Introduction}

Large language models (LLMs) have achieved remarkable results in natural language understanding and text generation through large-scale pre-training \cite{matarazzo2025survey}. However, relying solely on their internal knowledge often leads to hallucinations, outdated facts, and knowledge gaps. Retrieval-Augmented Generation (RAG) addresses this by integrating external search engines or knowledge bases into the generation process, grounding outputs in real-world evidence \cite{lewis2020goodrag}. Most RAG systems follow either a single-step “retrieve–reason” paradigm or a multi-step “plan–retrieve–reason” paradigm with a fixed retrieval plan \cite{lee2024planrag}. Although these methods improve factuality, they cannot adapt dynamically to intermediate findings, which can result in redundant searches and suboptimal use of retrieved documents.


Recent advances in reasoning-oriented LLMs, such as DeepSeek-R1 \cite{guo2025deepseekr1}, OpenAI-o1 \cite{openai2024o1}, and Qwen-QwQ \cite{qwen2025qwq32b}, have greatly enhanced chain-of-thought planning and tool invocation. Stronger reasoning capabilities enable precise decomposition of complex queries, dynamic adjustment of subsequent search steps based on partial evidence, and more effective synthesis and conflict resolution across multiple documents. This synergy has led to dynamic reasoning–retrieval model \& frameworks 
such as R1-Searcher \cite{song2025r1searcher}, Search-R1 \cite{jin2025searchr1}, and ReSearch \cite{research}, which interleave reasoning and search actions for more flexible information gathering.

However, existing dynamic retrieval–reasoning methods face two main challenges. \textbf{First}, unfiltered retrieval results often contain substantial irrelevant content, which can mislead the reasoning process—over 60\% of retrievals introduce errors \cite{wu2024clasheval}, and the utility of added context declines as prompt length grows \cite{tian2025relevance}, especially for smaller models \cite{fang2024small_bit_noise}. \textbf{Second}, appending full-text documents to the prompt dramatically increases token usage and computation cost, degrading inference efficiency and latency \cite{zhang2024mapreduce}. Addressing both noise suppression and inference efficiency is essential for truly effective dynamic retrieval reasoning.

To address these challenges, we propose OThink-SRR1, a novel reasoning-retrieval model \& framework, enhancing the iterative \textbf{S}earch, \textbf{R}efinement and \textbf{R}easoning (\textbf{OThink-SRR1}) capability of large reasoning models through reinforcement learning.
%
%
%
Specifically, once the retrieval results are obtained, they are immediately refined to extract the key information most relevant to the current context. 
The original documents are then removed so that subsequent reasoning focuses only on this refined content to stablish a robust foundation for subsequent reasoning.
We also design a reward function that provides rewards when the ground truth is found in the last retrieved document, its refined summary, and the final answer, and penalizes excessive retrievals. 
This encourages the model to generate more precise search queries, improve retrieval efficiency, and refine information to filter out noise, supporting more focused reasoning and leading to higher answer accuracy.
We train  OThink-SRR1 on Qwen2.5-7B-Instruct \& Qwen2.5-3B-Instruct and evaluate it on four standard multi-hop QA benchmarks. Experimental results show that OThink-SRR1 achieves higher answer accuracy and better retrieval efficiency than baseline methods. In multi-round settings, OThink-SRR1 also completes tasks with lower generation costs. In summary, we make the following contributions:




\begin{itemize}





    \item We propose the OThink-SRR1, which enhances the iterative Search-Refine-Reason capability of large reasoning models through reinforcement learning, establishing a foundational model for agents with information retrieval capabilities. To our best knowledge, OThink-SRR1 is the first large search reasoning model that internalizes and optimizes the ability to refine retrieval content through end-to-end training.

    \item We propose the RL algorithm GRPO-IR, which rewards accurate hits of the ground truth in the most recently retrieved documents, its refined summary, and the final response, while also penalizing excessive retrievals. GRPO-IR trains the large reasoning model to formulate more precise queries, enhance efficiency, and maintain focus on refined information.

    \item We systematically evaluate the OThink-SRR1 on multihop question answering benchmarks. Our model demonstrates superior performance compared to the latest baseline models in terms of EM/F1, while also decreasing the average retrieval count and total token usage.

    


\end{itemize}

\section{Related Work}

\subsection{Retrieval Augmented Generation}


Since being introduced, Retrieval-Augmented Generation (RAG) technology has seen substantial advancements. Early works focused on establishing an end-to-end framework for zero-shot QA \cite{du2022rag_eae}. With the growth in multimodal demands, the MuRAG system first achieved cross-modal knowledge fusion \cite{chen2022rag_murag}. Recent studies have made breakthroughs in dynamic retrieval strategies \cite{su2024rag_dragin}, long-tail knowledge augmentation \cite{li2024rag_longtail}, and noise-resistant training \cite{fang2024rag_enhancing}. Among these, the LongRAGE system improved long-text question-answering accuracy by 17\% through dual-perspective attention distillation \cite{zhao2024rag_longrag}, and the Provenance framework innovatively combined natural language inference models for factual verification of generated results \cite{sankararaman2024rag_provenance}.

As RAG frameworks continue to evolve to handle increasingly complex information needs, \textbf{multi-hop retrieval} has emerged as a crucial extension for reasoning across multiple pieces of evidence.  Early research focused on constraint query processing in knowledge graphs, such as the KG embedding method proposed in \cite{mitra2022multi_constraint}. The generative multi-hop retrieval model GMR \cite{lee2022multi_gen_multihop} optimized resource usage through text sequence generation. Dense-ATOMIC \cite{shen2023multi_dense} enhanced multi-hop path coverage by constructing dense knowledge graphs. 
Although RAG significantly expands the knowledge boundaries of LLMs, 
but current approaches primarily rely on single static retrievals, which limits LLMs' ability to handle complex tasks such as multi-hop reasoning.



\subsection{Retrieval for Reasoning in Large Language Models}


Recent research has explored various strategies to enhance the reasoning abilities of large language models (LLMs) by combining them with retrieval-augmented generation. 
Several works focus on modeling the retrieval process as a sequence of adaptive decisions, such as RAT \cite{wang2024rat}, which iteratively revises reasoning steps using relevant retrieved information to reduce hallucination. Other frameworks, including ReARTeR \cite{sun2025rearter} and AutoRAG \cite{yu2024autorag}, introduce methods like factuality scoring, process explanation, or autonomous multi-turn retrieval planning to further improve the integration of external information and the overall reliability of reasoning.

More recent approaches have adopted reinforcement learning (RL) to help LLMs actively learn when and how to interact with search engines during multi-step reasoning. For example, Search-R1 \cite{jin2025searchr1}, R1-Searcher \cite{song2025r1searcher}, and ReSearch \cite{research} present RL-based frameworks that allow LLMs to autonomously generate and refine search queries, select relevant information, and make step-wise decisions throughout the reasoning chain. These methods demonstrate that outcome-based rewards and the integration of retrieval actions within the reasoning process can significantly boost performance on complex question answering and knowledge-intensive tasks. However, current retrieval-reasoning models still suffer from issues such as excessive retrieval noise and high computational costs during inference.

\section{Method}

\subsection{Motivation}

Current RAG systems suffer from noisy retrievals and growing inefficiency. Irrelevant or incorrect documents can disrupt generation: over 60\% of retrievals introduce errors \cite{wu2024clasheval}, and relevance utility declines as context grows \cite{tian2025relevance}. Excessive content also distracts LLMs and wastes computation, degrading performance \cite{zhang2024mapreduce}. 
To address this, we introduce step‐wise reward for retrieval-reasoning models \cite{research,jin2025searchr1,song2025r1searcher}: after each retrieval, the LLM filters and retains only the most relevant documents, shortening context, boosting accuracy, and enhancing robustness.

\begin{figure*}[!tb]
    \centering
    \includegraphics[width=0.99\textwidth ]{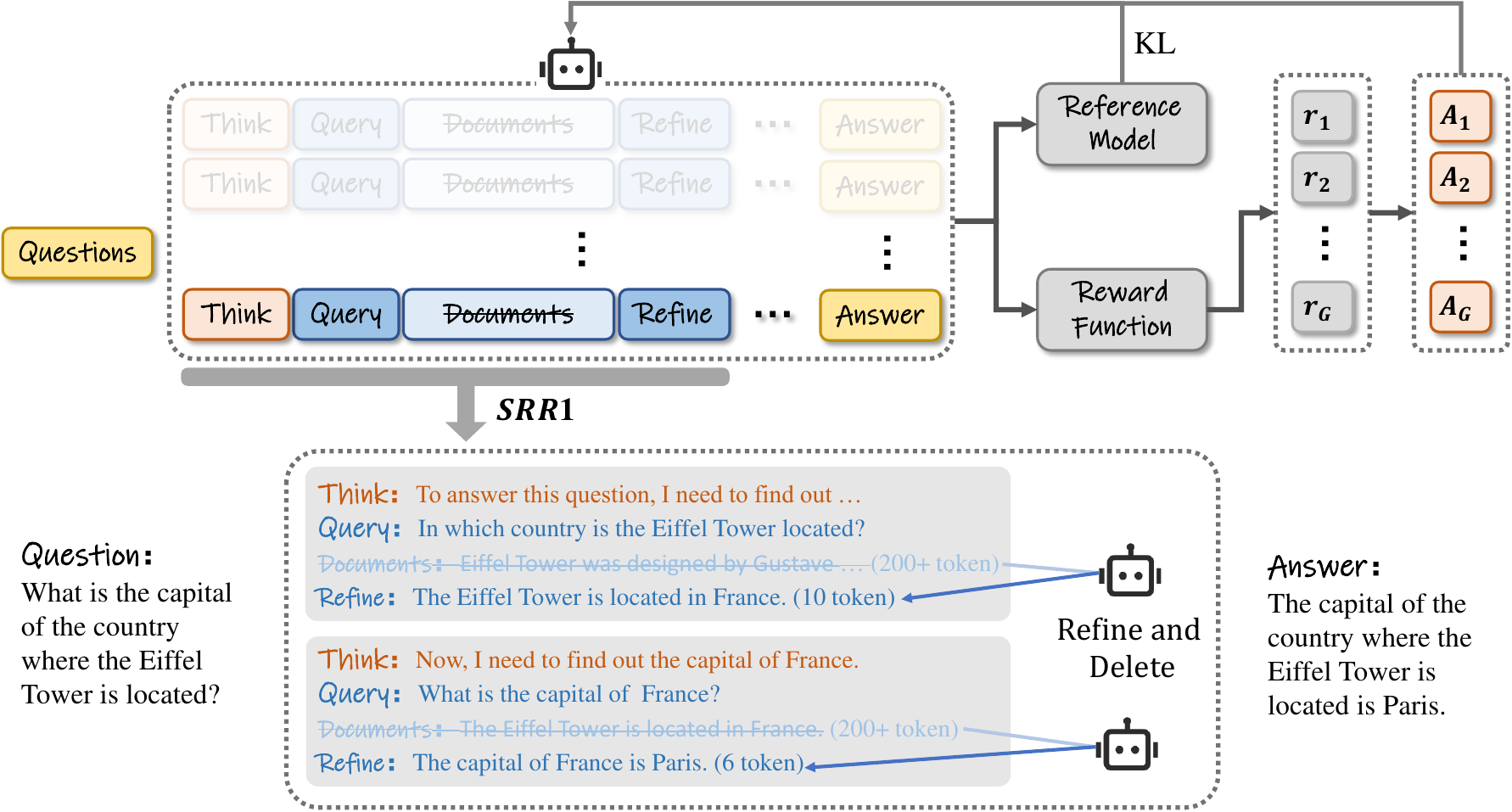} 
    \caption{Overview of our OThink-SRR1 framework. At each step the model (1) formulates a reasoning prompt ("Think"), (2) issues a retrieval query ("Query"), (3) condenses retrieved content into a concise answer ("Refine") to extract the key information most relevant to the current scenario. This process is iteratively executed until sufficient information is obtained to derive the final answer. Training is guided by rewards from a reference model and a stepwise reward function. 
    }
    \label{fig:overview}
\end{figure*}


\subsection{Search, Refine and Reasoning with Reinforcement Learning}


In this work, we introduce OThink-SRR1, a framework for multi-step Search-Refine-Reasoning integration in multi-hop retrieval tasks. Figure~\ref{fig:overview} provides an overview of our framework. Given a complex question $q$, our approach enables the model to autonomously develop search strategies, execute multiple rounds of querying with subsequent information refinement, and progressively accumulate essential knowledge until the final answer can be derived. 
%
%
Unlike other retrieval-reasoning methods \cite{song2025r1searcher,research,jin2025searchr1}, OThink-SRR1 distinguishes itself through two key features: (1) dynamic generation of high-quality retrieval queries during reasoning, and (2) a dedicated refinement step that systematically filters retrieved information to maximize relevance, which significantly improves the signal-to-noise ratio.
OThink-SRR1 requires models to demonstrate four core competencies: problem decomposition, retrieval strategy planning, information filtering, and evidence synthesis.


To achieve the aforementioned goals, we formulate the OThink-SRR1 task as a sequential decision-making problem, optimized via a reinforcement learning framework based on Group Relative Policy Optimization (GRPO).
Unlike the original GRPO, we have modified the reward mechanism in consideration of the task's characteristics: instead of relying on a reward model, we use an explicit reward function that is easier to control and measure. 
This approach not only reduces dependence on step-by-step human annotations or external reward models but also facilitates flexible adjustment of the reward strategy based on task requirements, thus efficiently driving the model to discover optimal solution strategies.

\subsubsection{Training System Prompt}

\definecolor{lightblue}{RGB}{235,245,255}
\definecolor{myblue}{RGB}{60,120,216}
\definecolor{mytitleblue}{RGB}{100,160,220}

\begin{figure*}[t]
\begin{tcolorbox}[
    colback=lightblue!80!white,       
    colframe=mytitleblue,     
    coltitle=white,      
    fonttitle=\bfseries, 
    title=System Prompt For Instruction-Tuned Model,
    colbacktitle=mytitleblue!90!white,  
    boxrule=1pt,         
    arc=2pt,             
    left=1mm, right=1mm, top=1mm, bottom=1mm, 
    label={prompt}, 
]
    You are a helpful assistant who employs a step-by-step approach using the Wikipedia search tool. Each step must be based ONLY on currently known information and lead to either a new search query or a final answer. Continue this sequential process until you have sufficient information to answer.

    \textbf{You must STRICTLY follow the output format below:}
    
    1. Place all reasoning, thinking, explanations, and process language INSIDE <think>...</think> tags.
    
    2. Place query INSIDE <query>...</query> tags for wikipedia search.
    
    3. After receiving search results (in <documents>...</documents>), based on currently known information, extract information relevant to the user question and current query INSIDE <documents\_refine>...</documents\_refine> tags.
    
    4. Place the final, direct, and concise answer ONLY INSIDE <answer>...</answer> tags, and wrap the direct, concise result WITHIN \textbackslash boxed\{\}.
    
    5. Do not output any text outside the <think>, <query>, <documents\_refine>, and <answer> tags.

    \textbf{For example:}
    
    <think>This is your thinking process.</think>
    <query>This is a search query for retrieval.</query>
    <documents>This is the search result.</documents>
    <documents\_refine>This is the refined information extracted from the documents that are most relevant to the query and enough to promote the next step.</documents\_refine>
    <think>Further reasoning, explanation or search.</think>
    <answer>The final answer is \textbackslash boxed\{exact answer here\}</answer>

\end{tcolorbox}
\caption{System Prompt for instruction-tuned model.}
\label{fig:prompt}
\end{figure*}

To ensure that the model consistently generates outputs in a well-defined structured format, we design prompts that explicitly specify the required output tags, as shown in Figure~\ref{fig:prompt}. In complex, multi-step reasoning tasks that require information retrieval, the model's internal thought process can be opaque, making it difficult to diagnose the root cause of errors. Conventional free-form text outputs exacerbate this challenge, as they are notoriously difficult to parse and evaluate automatically. Our structured format addresses this directly by requiring the model to externalize its reasoning into distinct, machine-readable components. Specifically, the model is instructed to use <query>...</query> tags to initiate retrieval operations and <documents\_refine>...</documents\_refine> tags to present refined information extracted from the retrieved documents. Furthermore, to enable precise Exact Match (EM) evaluation, the final answer is enclosed within a \textbackslash box\{\} environment. This design makes the final answer trivial to extract, eliminating ambiguity and facilitating robust, automated performance measurement.

This structured output format serves two main purposes. First, it allows for the automated evaluation of the model’s intermediate reasoning and retrieval steps, creating a transparent audit trail of its problem-solving process. By parsing the structured output, we can programmatically assess the quality of each step—for example, by evaluating the relevance of a <query> or the faithfulness of the information in <documents\_refine> to the source documents. This capability makes detailed error analysis and performance tracking possible, allowing us to move beyond simply knowing if a final answer is correct to understanding why the model succeeded or failed. This detailed insight is invaluable for targeted model improvement and debugging.

Second, this format provides clear and fine-grained feedback at different stages of the reasoning process, which is crucial for reinforcement learning (RL). In many complex tasks, a reward signal based solely on the final answer is often too sparse and delayed to be effective. Our approach allows for a more sophisticated, multi-faceted reward function that considers: format correctness, retrieval accuracy, refinement quality, final answer accuracy, and retrieval efficiency. This design enables the model to receive explicit and interpretable feedback for each discrete action it takes, such as formulating a query or synthesizing information. By providing immediate rewards for successful intermediate steps, we can guide the model toward more effective reasoning strategies and systematically improve its performance on these critical sub-tasks.

\subsubsection{GRPO‐IR}

Multi‐hop retrieval tasks require models to chain together information from several documents, but intermediate retrieval and reasoning steps usually lack direct supervision. To address this, we adapt Group Relative Policy Optimization (GRPO) \cite{shao2024deepseekmath} by inserting a non‐differentiable information retrieval (IR) operation into the autoregressive loop.

Concretely, let $q\sim \mathcal{Q}$ be an input question drawn from distribution $\mathcal{Q}$. At each dedicated ``\texttt{<query>}'' token, the policy $\pi_{\theta}$ issues a search to an external retriever and appends the top-$K$ documents to the generation context. These retrieved tokens are treated as non-differentiable during training: we block gradients through them and exclude them from importance-weight ratios and KL computations.

We then update $\pi_{\theta}$ by maximizing the following clipped, KL‐regularized objective over groups of $G$ sampled trajectories $\{a_i\}_{i=1}^G$:


{\footnotesize
\begin{equation}
\label{grpo}
\begin{aligned}
 \mathcal{J}&(\theta) =\mathbb{E}_{q \sim \mathcal{Q},\left\{a_{i}\right\}_{i=1}^{G} \sim \pi_{\theta_{\text {old }}}(\cdot \mid q)} \\ 
    & \frac{1}{G} \sum_{i=1}^{G} \Biggl[ \min \Biggl(\frac{\pi_{\theta}\left(a_{i} \mid q;IR\right)}{\pi_{\theta_{\text {old }}}\left(a_{i} \mid q;{IR}\right)} A_{i}, \\ 
    & \operatorname{clip}\left(\frac{\pi_{\theta}\left(a_{i} \mid q ;{IR}\right)}{\pi_{\theta_{\text {old }}}\left(a_{i} \mid q ;{IR} \right) }, 1-\epsilon, 1+\epsilon\right) A_{i} \Biggr) \\
    & -\beta \mathbb{D}_{\text {KL }}\left(\pi_{\theta}| | \pi_{\theta_{\text {ref }}}\right) \Biggr]
\end{aligned}
\end{equation}
}

Each $a_i$ is a full generated sequence, including reasoning tokens, query tokens, retrieved documents, and refinements and $\mathbf{r}=\{r_i\}_{i=1}^G$ are the rewards computed by our hierarchical reward function. Here,
$\theta$ is the current policy, $\theta_{\rm old}$ the policy at the start of the update, and $\theta_{\rm ref}$ a fixed ``reference'' policy used for KL regularization, $\epsilon$ is the clipping parameter (e.g., $0.2$), $\beta$ controls the weight of the KL penalty, $\beta$ controls the weight of the KL penalty, $K$ is the number of top documents retrieved at each query step, $G$ is the group size (number of trajectories) used to compute relative advantages, and $A_{i}=\widetilde{r}_{i}=\frac{r_{i}-\operatorname{mean}(\mathbf{r})}{\operatorname{std}(\mathbf{r})}$.
%

In practice, for each $q$ we (1) sample $G$ full sequences under $\pi_{\theta_{\rm old}}$, (2) compute per‐sequence rewards $r_i$ with our reward function (see Section~\ref{sec:reward}), (3) standardize them to obtain advantages $A_i$, and (4) take a gradient step on $\mathcal{J}(\theta)$. 

\subsubsection{Reward Function}
\label{sec:reward}

We design a hierarchical and interpretable reward function that decomposes the complex reasoning task into distinct sub-goals: format correctness, retrieval accuracy, and answer relevance. The total reward is structured as a sum of bonuses awarded for successfully completing each step.

First, outputs must strictly adhere to the required format (e.g., <think>...</think>). Failure to do so results in a reward of zero, enforcing structural integrity. For correctly formatted outputs, the reward calculation proceeds as follows: The model receives a base reward ($\beta_{base}$) for any valid attempt. A significant retrieval bonus ($\beta_{retrieval}$) is added if the last retrieved document contains the ground truth (GT). Crucially, to encourage the model to not only find but also correctly utilize the information, an additional refinement bonus ($\beta_{refine}$) is granted if the last refined content also contains the GT. This layered bonus system explicitly incentivizes effective information synthesis beyond simple retrieval.

Alongside correctness, the reward function also incorporates retrieval efficiency, scored by $r_{count}$. This component encourages the model to minimize the number of retrievals. We define a maximum efficiency reward, $\eta_{max}$, awarded when the number of retrieval steps $C$ is at or below an ideal threshold $C_{start}$. Beyond this threshold, the reward linearly decays to a minimum efficiency reward, $\eta_{min}$, as $C$ approaches the upper limit $C_{max}$.

The overall reward $r$ is then determined by combining these components. We define three indicator variables: 

\begin{itemize}
    \item \textit{F}: equals 1 if the output format is correct, 0 otherwise.
    \item \textit{D}: equals 1 if the last retrieved document contains the ground truth (GT), 0 otherwise.
    \item \textit{R}: equals 1 if the last refined content contains GT, 0 otherwise.
\end{itemize}

\begin{equation}
\resizebox{0.9\columnwidth}{!}{%
$\displaystyle
    r = 
    \begin{cases} 
    \beta_{base} + \beta_{retrieval} + \beta_{refine} + \text{f1} + r_{\text{count}}, & \text{if }F=D=R=1 \\
    \beta_{base} + \beta_{retrieval} + r_{\text{count}}, & \text{if }F=D=1,R=0 \\
    \beta_{base}, & \text{if }F=1,D=0 \\
    0, & \text{if }F=0 \\
    \end{cases}
  $
}
\end{equation}

Here, f1 denotes the F1 score between the model's final answer and the GT, computed as:
\begin{equation}
\text{f1} = \frac{2 \times m}{p + q}
\end{equation}
where $p$ and $q$ are the word counts of the predicted and reference answers, respectively, and $m$ is the number of overlapping words between the two. This metric captures the similarity between the two answers by considering both precision and recall. The efficiency score $r_{count}$ is defined as:
\begin{equation}
\resizebox{0.9\columnwidth}{!}{%
$\displaystyle
  r_{count} = 
  \begin{cases} 
  \eta_{max}, & \text{if } C \leq C_{start} \\
  \eta_{min} + (\eta_{max} - \eta_{min}) \frac{C_{max} - C}{C_{max} - C_{start}}, & \text{if } C_{start} < C < C_{max} \\
  \eta_{min}, & \text{if } C \geq C_{max} \\
  \end{cases}
  $
}
\end{equation}

In our experiments, the reward hyperparameters are set to $\beta_{base}=0.1$, $\beta_{retrieval}=0.2$, $\beta_{refine}=0.3$, $\eta_{max}=0.2$, and $\eta_{min}=0$. This configuration creates a clear reward hierarchy that prioritizes correct refinement over simple retrieval, and concise reasoning over excessive steps. The modular design also allows for future work to systematically tune the relative importance of each reasoning component.

The reward function thus considers the accuracy of the final answer, the efficiency in retrieval steps, and the format correctness, promoting both effectiveness and efficiency in the model's generation process.

\section{Experiments}

\subsection{Experimental Setting}

To evaluate the effectiveness of OThink-SSR1, we conducted extensive experiments on multi-hop QA benchmarks that require multiple information retrievals. Our OThink-SSR1 models were trained using Qwen2.5-7B-Instruct and Qwen2.5-3B-Instruct. During the training process, we exclusively utilized data from the MuSiQue training set.

\textbf{Benchmarks} 
%
We evaluated on four multi-hop QA benchmarks: HotpotQA \cite{yang2018hotpotqa}, 2WikiMultiHopQA \cite{ho20202Wikimultihopqa}, MuSiQue \cite{trivedi2022musique}, and Bamboogle \cite{press2023bamboogle}. The first three were built from Wikipedia/Wikidata \cite{wikidata} using different multi-hop strategies, while Bamboogle contains challenging 2-hop questions. We used development sets (HotpotQA: 7405; 2WikiMultiHopQA: 12576; MuSiQue: 2417) and Bamboogle's test set (125 samples), removing original contexts and relying solely on Wikipedia retrieval for background knowledge.

\textbf{Baselines} 
%
We compared OThink-SRR1 with several baselines: (1) No RAG (direct generation); (2) Basic RAG: simple retrieval-augmented generation; (3) Iter-RetGen \cite{shao2023iterretgen}: iterative retrieval-generation; and (4) IRCoT \cite{trivedi2023ircot}: interleaved retrieval and Chain-of-Thought reasoning. To ensure a fair comparison, all methods were based on the same family of instruction-tuned models. Specifically, the Search-R1\cite{jin2025searchr1} and the 7B-Instruct version of the ReSearch\cite{research} baseline leveraged publicly available open-source models. For ReSearch's 3B-Instruct variant, we trained the model using its official implementation with the same hyperparameters as OThink-SRR1 to create a controlled comparison.

\textbf{Evaluation Metrics} To assess the correctness of final answers, we evaluated answer quality using: (1) Exact Match (EM) for strict correctness, and (2) f1 score for partial matches at token level. Both metrics are reported to comprehensively assess performance.


\textbf{Implementation Details} 
%
Experiments ran on 4 × A100-80GB GPUs, training used batch size 256 for 2 epochs with 5e-6 learning rate. Inference applied sampling (temp=0.7, top$\_$p=0.9) with 8192-token limit. All methods shared the same ElasticSearch-based Wikipedia retrieval (6M articles, 2023 version), fetching top-5 passages, the maximum retrieval limit $C_{max}=5$.

\subsection{Results Analysis}

In this section, we report and analyze the effectiveness of the proposed method (OThink-SRR1) across various settings and benchmarks, focusing on its performance and generalization capabilities as evidenced by Table~\ref{tab:conbined}.

\paragraph{OThink-SRR1 demonstrates superior performance on multi-hop QA benchmarks.}
As detailed in Table~\ref{tab:conbined}, OThink-SRR1 consistently outperforms established baselines across most evaluated datasets when using Qwen2.5-7B-Instruct. It achieves an average Exact Match (EM) of 36.26\% and an F1 score of 45.85\%, surpassing the strong ReSearch baseline by 0.91 EM and 0.27 F1 points. Notably, on the complex MuSiQue dataset, OThink-SRR1 improves EM to 20.73\% from ReSearch's 19.07\%.  These results underscore the efficacy of our orchestrated search-retrieve-reasoning framework. This performance superiority is also maintained with the smaller Qwen2.5-3B-Instruct model, where OThink-SRR1 achieves an average EM of 28.61\%, significantly ahead of other methods.

\paragraph{OThink-SRR1 exhibits strong generalization to out-of-domain datasets.}
Despite being trained exclusively on the MuSiQue dataset, OThink-SRR1 shows robust generalization to unseen benchmarks such as 2WikiMultiHopQA and Bamboogle. On these out-of-domain datasets, our method still achieves leading performance, with the 7B model attaining 41.52\% EM on 2WikiMultiHopQA and 44.00\% EM on Bamboogle, exceeding ReSearch. This effective transfer suggests that OThink-SRR1 learns fundamental and adaptable reasoning and retrieval strategies rather than overfitting to the characteristics of the training data. The generalization capability is also evident with the 3B model, further highlighting the robustness of our approach across different model scales and data distributions.

\begin{table*}[htbp]
  \small
  \centering
  \caption{The results of different methods on four multi-hop QA datasets, bolded result represent better outcomes for correspond model parameters and metrics.}
  \label{tab:conbined}
  \setlength{\tabcolsep}{4.5pt}
  \begin{tabular*}{\textwidth}{@{\extracolsep{\fill}}l cc|cc|cc|cc|cc@{}}
    \toprule
    \multirow{2}{*}{\textbf{Model}}
      & \multicolumn{2}{c}{\textbf{MuSiQue}}
      & \multicolumn{2}{c}{\textbf{2Wiki}}
      & \multicolumn{2}{c}{\textbf{Bamboogle}}
      & \multicolumn{2}{c}{\textbf{HotpotQA}} 
      & \multicolumn{2}{c}{\textbf{Avg.}} \\
    \cmidrule(lr){2-3} \cmidrule(lr){4-5} \cmidrule(lr){6-7} \cmidrule(lr){8-9} \cmidrule(lr){10-11}
      & EM   & F1   & EM    & F1    & EM    & F1    & EM     & F1     & EM $\uparrow$     & F1 $\uparrow$ \\
    \midrule
    \multicolumn{11}{c}{\textbf{Base model: Qwen2.5-7B-Instruct}} \\
No RAG
      & 3.60 & 11.59 & 25.52 & 30.32 & 10.40 & 17.57 & 19.41 & 27.50 & 14.73 & 21.75 \\
Basic RAG
      & 6.45 & 13.09 & 25.59 & 32.01 & 20.00 & 29.34 & 32.13 & 42.16 & 21.04 & 29.15 \\
Iter-RetGen
      & 8.19 & 15.79 & 27.85 & 34.17 & 20.00 & 29.32 & 34.64 & 45.40 & 22.67 & 31.17 \\
IRCOT
      & 6.83 & 13.50 & 21.49 & 29.28 & 22.40 & 34.54 & 30.61 & 42.20 & 20.33 & 29.88 \\
Search-R1
      & 16.88 & 24.58 & 33.08 & 39.03 & 39.20 & 49.71 & 38.24 & 48.50 & 31.85 & 40.45 \\
ReSearch
      & 19.07 & {28.66} & 40.6 & 48.66 & 42.40 & 53.65 & \textbf{39.32} & \textbf{51.35} & 35.35 & 45.58 \\
      \midrule
OThink-SRR1
      & \textbf{20.73} & \textbf{29.85} & \textbf{41.52} & \textbf{49.33} & \textbf{44.00} & \textbf{53.69} & 38.80 & 50.55 & \textbf{36.26} & \textbf{45.85} \\
      \midrule
     \multicolumn{11}{c}{\textbf{Base model: Qwen2.5-3B-Instruct}} \\
No RAG
      & 2.36 & 7.92 & 24.79 & 28.73 & 2.40 & 8.94 & 16.02 & 22.43 & 11.39 & 17.00 \\
Basic RAG
      
      & 5.63 & 10.94 & 24.54 & 30.18 & 9.60 & 19.06 & 28.10 & 37.45 & 16.97 & 24.41 \\
Iter-RetGen
      
      & 7.12 & 12.61 & 25.97 & 31.47 & 11.20 & 19.09 & 30.09 & 39.56 & 18.59 & 25.68 \\
IRCOT
      & 6.79 & 12.51 & 21.45 & 30.00 & 20.80 & 31.47 & 27.20 & 37.50 & 19.06 & 27.87 \\
Search-R1
      
      & 10.26 & 17.05 & 32.11 & 37.87 & 28.00 & 38.12 & 30.20 & 39.62 & 25.14 & 33.16 \\
ReSearch
      & 12.45 & 19.32 & 26.82 & 32.22 & 26.40 & 36.45 & 30.11 & 39.91 & 23.95 & 31.98\\
      
\midrule
OThink-SRR1
      & \textbf{14.98} & \textbf{23.94} & \textbf{35.27} & \textbf{43.54} & \textbf{31.20} & \textbf{43.58} & \textbf{32.99} & \textbf{44.21} & \textbf{28.61} & \textbf{38.82} \\
    \bottomrule
  \end{tabular*}
\end{table*}

\subsection{Further Analysis}

\begin{figure*}[!tb]
    \centering
    \includegraphics[width=0.99\textwidth ]{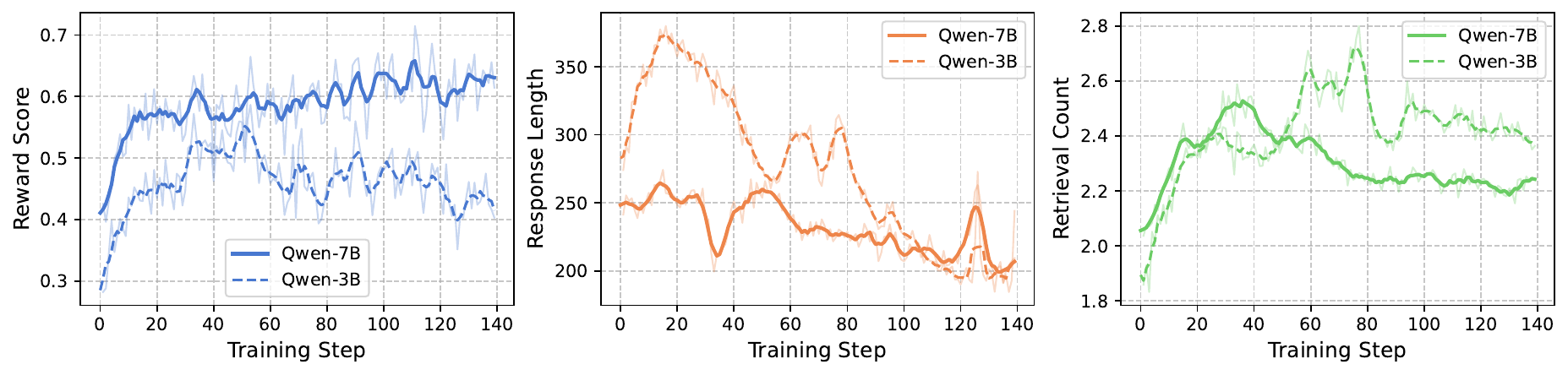} 
    \caption{
    %
    The reward scores, response lengths, and retrieval counts during the training process of Qwen2.5-3B-Instruct and Qwen2.5-7B-Instruct models.
    }
    \label{fig:line_diagram}
\end{figure*}


\textbf{Training Dynamics Analysis.}
Figure~\ref{fig:line_diagram} plots reward score, response length, and retrieval count for both Qwen-7B (solid) and Qwen-3B (dashed) under OThink-SRR1. Both models exhibit an early exploration phase—retrieval count rise to about 2.5–2.8 per question, followed by consolidation-retrieval count fall to 2.2-2.4 and responses shorten by roughly 25\%. Simultaneously, rewards climb from 0.4 to 0.6 for 7B and from 0.3 to 0.5 for 3B before plateauing, demonstrating that OThink-SRR1 learns to filter noise and focus on key information. This transition from broad search to efficient reasoning highlights OThink-SRR1’s strength in reducing redundant retrievals and context size—a critical advantage for smaller models prone to distraction by noisy inputs.

\begin{figure*}[!tb]
    \centering
    \includegraphics[width=0.99\textwidth ]{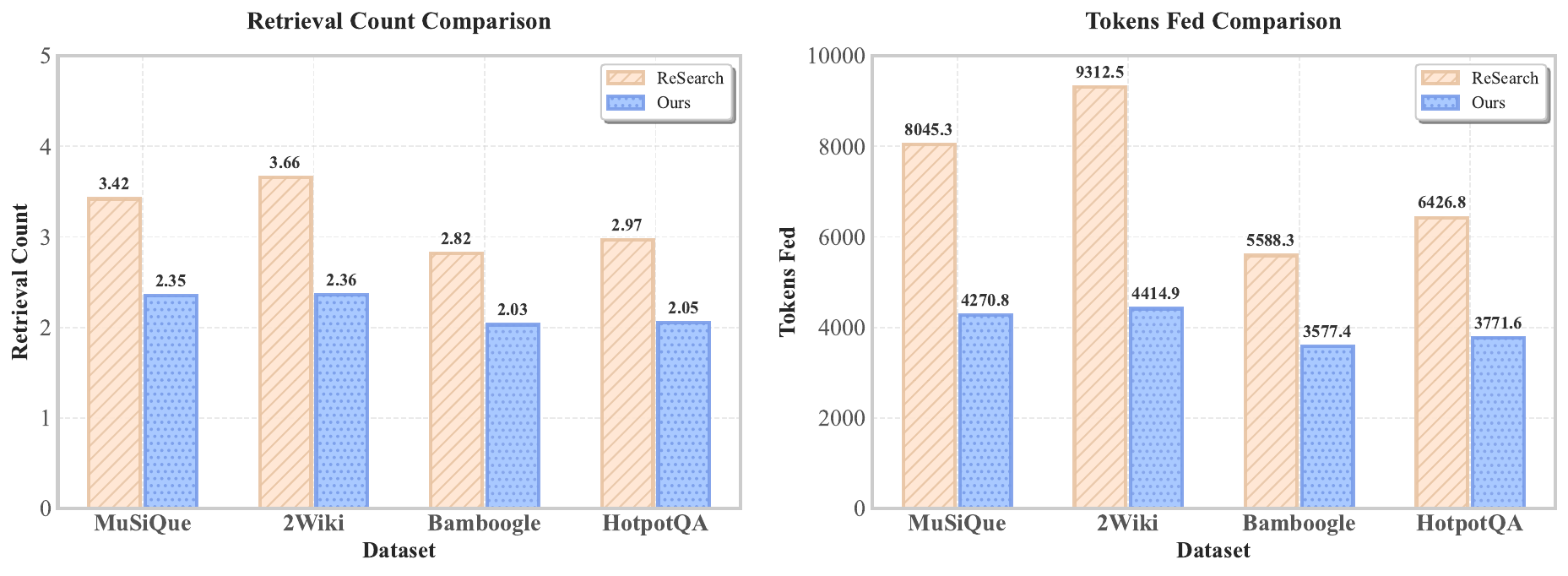} 
    \caption{Comparison of Average Retrieval Count and Total Token Usage Between OThink-SRR1 and the Baseline ReSearch on Multi-hop QA Datasets, based on Qwen2.5-7B-Instruct model.
    }
    \label{fig:bar_diagram}
\end{figure*}

\definecolor{qg_background}{RGB}{245,245,245}         
\definecolor{our_method_background}{RGB}{230,240,255} 
\definecolor{comp_method_background}{RGB}{230,250,230}
\definecolor{mytitleblue}{RGB}{100,160,220}


\begin{figure*}[!h]
\begin{tcolorbox}[
    fontupper=\footnotesize,
    colback=white,
    colframe=mytitleblue,
    width=\textwidth,
    arc=2pt,
    boxrule=0.8pt,
    left=0mm, right=0mm, top=0mm, bottom=0mm,
    boxsep=0pt,
    enhanced,
]

    \begin{tcolorbox}[
        colback=qg_background,
        boxrule=0pt,
        arc=0pt,
        sharp corners,
        left=2mm, right=2mm, top=1mm, bottom=1mm,
        boxsep=0pt, width=\textwidth, enlarge left by=0mm, enlarge right by=0mm,
        nobeforeafter
    ]

    \textbf{Question:} What was the population of the city that is found in the same county as Eric Marcus Municipal Airport in 1900? \\
    \textbf{Ground Truth:} 7,531
    \end{tcolorbox}%

    \begin{tcolorbox}[
        colback=comp_method_background,
        boxrule=0pt,
        arc=0pt,
        sharp corners,
        left=2mm, right=2mm, top=1mm, bottom=1mm,
        boxsep=0pt, width=\textwidth, enlarge left by=0mm, enlarge right by=0mm,
        nobeforeafter
    ]
    \textbf{ReSearch (7B) Thinking Process:} \hfill {\textbf{{tokens fed to inference: 10347}}} 

    \texttt{<think>  To find ... planing ... </think>,}\\
    \textcolor{red}{<search> What county is Eric Marcus Municipal Airport in? </search>},\\
    \texttt{<result> ... Eric Marcus Municipal Airport is a county-owned, public-use airport in Pima County, Arizona, ... </result>,}\\
    \texttt{<think> ... I will now find the city in that county ...</think>,}\\
    \textcolor{red}{<search> What is the largest city in Pima County, Arizona? </search>,}\\
    \texttt{<result> ... Pima County includes the Tucson, ... </result>,}\\
    \texttt{<think> ... I will find the population of Tucson in the year 1900 ...</think>,}\\
    \textcolor{red}{<search> What was the population of Tucson, Arizona in 1900? </search>,}\\
    \texttt{<result> ... By 1900, 7,531 people lived in the city ... </result>,}\\
    \texttt{<think> After finding the population of Tucson in 1900, I can now provide the answer to the question.</think>,}\\
    \textcolor{red}{<search> What was the population of Tucson, Arizona in 1900? </search>,}\\
    \texttt{<result> ... By 1900, 7,531 people lived in the city ... </result>,}\\
    \texttt{<think> Having found the population of Tucson in 1900, I can now provide the final answer to the question.</think>,}\\
    \texttt{<answer> \textbackslash{}boxed\{Tucson\} </answer>}
    \end{tcolorbox}

    \begin{tcolorbox}[
        colback=our_method_background,
        boxrule=0pt,
        arc=0pt,
        sharp corners,
        left=2mm, right=2mm, top=1mm, bottom=1mm,
        boxsep=0pt, width=\textwidth, enlarge left by=0mm, enlarge right by=0mm,
        nobeforeafter
    ]
    \textbf{OThink-SRR1-7B Thinking Process:} \hfill {\textbf{{tokens fed to inference: 3638}}}\\
    \texttt{<think> First, ... identify the ... Airport  is located. </think>,}\\
    \textcolor{red}{<query> What county is Eric Marcus Municipal Airport in?</query>}\\
    \sout{<documents> ...airport in Pima County, Arizona,... </documents>}\\
    \textbf{<documents\_refine> Pima County, Arizona. </documents\_refine>,}\\
    \texttt{<think> ...find the city in that county and its 1900 population. </think>}\\
    \textcolor{red}{<query> What city in Pima County, Arizona had a population in 1900?</query>}\\
    \sout{<documents> ...By 1900, 7,531 people lived in the city... </documents>}\\
    \textbf{<documents\_refine> Tucson, Arizona had a population of 7,531 in 1900. </documents\_refine>}\\
    \texttt{<think> The city that was in the same county as Eric Marcus Municipal Airport in 1900 was Tucson, and its population in 1900 is found to be 7,531. </answer>}\\
    \texttt{<answer> \textbackslash{}boxed\{7531\} </answer>}
    \end{tcolorbox}%

\end{tcolorbox}


    \caption{Case Study Comparison Between OThink-SRR1 and ReSearch Baseline. 
\textcolor{red}{Red} queries show search operations, \sout{strikethrough} shows discarded documents. }


    \label{fig:reasoning-case}
\end{figure*}


Figure~\ref{fig:bar_diagram} provides a detailed comparison of average retrieval counts and cumulative token consumption between our proposed \textbf{ OThink-SRR1} framework and the baseline ReSearch across four representative multi-hop QA datasets. 

As shown in the left panel, our method consistently achieves a lower average retrieval count (2.35) compared to the baseline (3.42) on all datasets, indicating that \textbf{ OThink-SRR1} is able to reduce unnecessary or redundant retrieval steps during the reasoning process. Correspondingly, the right panel demonstrates a significant reduction in the total number of input tokens processed by the model (4270.81 vs.\ 8045.26), reflecting improved computational efficiency and lower resource consumption. This reduction in both retrieval count and token usage not only alleviates the burden on the model but also contributes to more focused and effective reasoning. Importantly, these efficiency gains do not come at the cost of performance; instead, our method achieves stronger results, validating that the Search-Refine-Reasoning paradigm enables the model to filter out noise and concentrate on relevant information, leading to both better accuracy and improved efficiency.


\subsection{Case Study}

The case study (Figure~\ref{fig:reasoning-case}) vividly illustrates OThink-SRR1's practical superiority. OThink-SRR1 efficiently solves the multi-hop query in just two retrieval-refinement cycles, processing only 3,638 tokens. Each refinement cycle acts as a powerful distillation step, transforming verbose documents into compact, factual assertions. By iteratively retaining only these essential facts (e.g., "Pima County, Arizona," then "Tucson...population of 7,531"), it prevents the reasoning context from being cluttered. This strategic pruning is not merely about efficiency; it critically reduces the model's cognitive load, allowing it to maintain focus and correctly extract the population "7,531". In stark contrast, the ReSearch baseline, lacking this refinement, processes significantly more tokens (10,347) across more steps. It becomes overwhelmed by the noisy, unpruned context, leading to a classic failure of contextual distraction where a salient but incorrect entity ("Tucson") hijacks the model's attention. This comparison demonstrates that for complex tasks, the ability to intelligently filter and forget is as vital as the ability to retrieve and reason. OThink-SRR1's success proves that effective reasoning is not just about finding information, but about building a clear path to the answer by judiciously discarding noise.


\section{Conclusion}

We introduced OThink-SRR1, a novel framework that synergistically integrates search, refinement, and R1-style reasoning through reinforcement learning, specifically targeting retrieval noise and inefficiency in multi-hop question answering. 
Our GRPO-IR reward mechanism enables dynamic information management without supervised intermediate steps.
Extensive experiments demonstrate OThink-SRR1's significant outperformance against strong baselines across four benchmarks in terms of accuracy, retrieval efficiency, and token economy. The framework also exhibits strong generalization capabilities from single-dataset training, offering a robust and efficient approach to advance LLM performance in complex, knowledge-intensive reasoning tasks.



\clearpage

\section{Limitation}


Despite its effectiveness, OThink-SRR1's refinement process occasionally discards critical information needed for accurate reasoning. This loss of essential details can lead to incorrect answers, particularly when relevant context is subtle or dispersed across documents. 
Future work will enhance context-awareness through adaptive filtering to preserve key information while reducing noise. Incorporating dynamic reward functions based on task-specific feedback could improve flexibility. These refinements aim to broaden OThink-SRR1's applicability to complex reasoning tasks.

%

\bibliography{Search}

@article{wu2024clasheval,
  title={Clasheval: Quantifying the tug-of-war between an llm’s internal prior and external evidence},
  author={Wu, Kevin and Wu, Eric and Zou, James Y},
  journal={Advances in Neural Information Processing Systems},
  volume={37},
  pages={33402--33422},
  year={2024}
}

@inproceedings{tian2025relevance,
  title={Is Relevance Propagated from Retriever to Generator in RAG?},
  author={Tian, Fangzheng and Ganguly, Debasis and Macdonald, Craig},
  booktitle={European Conference on Information Retrieval},
  pages={32--48},
  year={2025},
  organization={Springer}
}

@article{zhang2024mapreduce,
  title={A MapReduce Approach to Effectively Utilize Long Context Information in Retrieval Augmented Language Models},
  author={Zhang, Gongbo and Xu, Zihan and Jin, Qiao and Chen, Fangyi and Fang, Yilu and Liu, Yi and Rousseau, Justin F and Xu, Ziyang and Lu, Zhiyong and Weng, Chunhua and others},
  journal={arXiv preprint arXiv:2412.15271},
  year={2024}
}

@inproceedings{fang2024small_bit_noise,
    title = "Enhancing Noise Robustness of Retrieval-Augmented Language Models with Adaptive Adversarial Training",
    author = "Fang, Feiteng  and
      Bai, Yuelin  and
      Ni, Shiwen  and
      Yang, Min  and
      Chen, Xiaojun  and
      Xu, Ruifeng",
    editor = "Ku, Lun-Wei  and
      Martins, Andre  and
      Srikumar, Vivek",
    booktitle = "Proceedings of the 62nd Annual Meeting of the Association for Computational Linguistics (Volume 1: Long Papers)",
    month = aug,
    year = "2024",
    address = "Bangkok, Thailand",
    publisher = "Association for Computational Linguistics",
    url = "https://aclanthology.org/2024.acl-long.540/",
    doi = "10.18653/v1/2024.acl-long.540",
    pages = "10028--10039",
}

@article{matarazzo2025survey,
  title={A Survey on Large Language Models with some Insights on their Capabilities and Limitations},
  author={Matarazzo, Andrea and Torlone, Riccardo},
  journal={arXiv preprint arXiv:2501.04040},
  year={2025}
}

@article{guo2025deepseekr1,
  title={Deepseek-r1: Incentivizing reasoning capability in llms via reinforcement learning},
  author={Guo, Daya and Yang, Dejian and Zhang, Haowei and Song, Junxiao and Zhang, Ruoyu and Xu, Runxin and Zhu, Qihao and Ma, Shirong and Wang, Peiyi and Bi, Xiao and others},
  journal={arXiv preprint arXiv:2501.12948},
  year={2025}
}

@article{shao2024deepseekmath,
  title={Deepseekmath: Pushing the limits of mathematical reasoning in open language models},
  author={Shao, Zhihong and Wang, Peiyi and Zhu, Qihao and Xu, Runxin and Song, Junxiao and Bi, Xiao and Zhang, Haowei and Zhang, Mingchuan and Li, YK and Wu, Y and others},
  journal={arXiv preprint arXiv:2402.03300},
  year={2024}
}

@misc{openai2024o1,
  author = {{OpenAI}},
  title = {Learning to Reason with LLMs},
  year = {2024},
  howpublished = {\url{https://openai.com/index/learning-to-reason-with-llms/}},
  note = {Accessed: 2024-06-13}
}

@misc{qwen2025qwq32b,
    title = {QwQ-32B: Embracing the Power of Reinforcement Learning},
    author = {{Qwen Team}},
    month = {March},
    year = {2025},
    howpublished = {\url{https://qwenlm.github.io/blog/qwq-32b/}},
    note = {Accessed: 2025-03-6}
}

@article{research,
  title={Learning to Reason with Search for LLMs via Reinforcement Learning},
  author={Chen, Mingyang and Li, Tianpeng and Sun, Haoze and Zhou, Yijie and Zhu, Chenzheng and Yang, Fan and Zhou, Zenan and Chen, Weipeng and Wang, Haofen and Pan, Jeff Z and others},
  journal={arXiv preprint arXiv:2503.19470},
  year={2025}
}

@article{jin2025searchr1,
  title={Search-r1: Training llms to reason and leverage search engines with reinforcement learning},
  author={Jin, Bowen and Zeng, Hansi and Yue, Zhenrui and Wang, Dong and Zamani, Hamed and Han, Jiawei},
  journal={arXiv preprint arXiv:2503.09516},
  year={2025}
}

@article{song2025r1searcher,
  title={R1-Searcher: Incentivizing the Search Capability in LLMs via Reinforcement Learning},
  author={Song, Huatong and Jiang, Jinhao and Min, Yingqian and Chen, Jie and Chen, Zhipeng and Zhao, Wayne Xin and Fang, Lei and Wen, Ji-Rong},
  journal={arXiv preprint arXiv:2503.05592},
  year={2025}
}

@inproceedings{lee2024planrag,
  title={PlanRAG: A plan-then-retrieval augmented generation for generative large language models as decision makers},
  author={Lee, Myeonghwa and An, Seonho and Kim, Min-Soo},
  booktitle={Proceedings of the 2024 Conference of the North American Chapter of the Association for Computational Linguistics: Human Language Technologies (Volume 1: Long Papers)},
  pages={6537--6555},
  year={2024}
}

@inproceedings{shao2023iterretgen,
    title = "Enhancing Retrieval-Augmented Large Language Models with Iterative Retrieval-Generation Synergy",
    author = "Shao, Zhihong  and
      Gong, Yeyun  and
      Shen, Yelong  and
      Huang, Minlie  and
      Duan, Nan  and
      Chen, Weizhu",
    editor = "Bouamor, Houda  and
      Pino, Juan  and
      Bali, Kalika",
    booktitle = "Findings of the Association for Computational Linguistics: EMNLP 2023",
    month = dec,
    year = "2023",
    address = "Singapore",
    publisher = "Association for Computational Linguistics",
    url = "https://aclanthology.org/2023.findings-emnlp.620/",
    doi = "10.18653/v1/2023.findings-emnlp.620",
    pages = "9248--9274",
}

@inproceedings{trivedi2023ircot,
    title = "Interleaving Retrieval with Chain-of-Thought Reasoning for Knowledge-Intensive Multi-Step Questions",
    author = "Trivedi, Harsh  and
      Balasubramanian, Niranjan  and
      Khot, Tushar  and
      Sabharwal, Ashish",
    editor = "Rogers, Anna  and
      Boyd-Graber, Jordan  and
      Okazaki, Naoaki",
    booktitle = "Proceedings of the 61st Annual Meeting of the Association for Computational Linguistics (Volume 1: Long Papers)",
    month = jul,
    year = "2023",
    address = "Toronto, Canada",
    publisher = "Association for Computational Linguistics",
    url = "https://aclanthology.org/2023.acl-long.557/",
    doi = "10.18653/v1/2023.acl-long.557",
    pages = "10014--10037",
}

@article{wang2024rat,
  title={Rat: Retrieval augmented thoughts elicit context-aware reasoning in long-horizon generation},
  author={Wang, Zihao and Liu, Anji and Lin, Haowei and Li, Jiaqi and Ma, Xiaojian and Liang, Yitao},
  journal={arXiv preprint arXiv:2403.05313},
  year={2024}
}

@article{sun2025rearter,
  title={ReARTeR: Retrieval-Augmented Reasoning with Trustworthy Process Rewarding},
  author={Sun, Zhongxiang and Wang, Qipeng and Yu, Weijie and Zang, Xiaoxue and Zheng, Kai and Xu, Jun and Zhang, Xiao and Yang, Song and Li, Han},
  journal={arXiv preprint arXiv:2501.07861},
  year={2025}
}

@article{yu2024autorag,
  title={Auto-rag: Autonomous retrieval-augmented generation for large language models},
  author={Yu, Tian and Zhang, Shaolei and Feng, Yang},
  journal={arXiv preprint arXiv:2411.19443},
  year={2024}
}

@inproceedings{yang2018hotpotqa,
    title = "{H}otpot{QA}: A Dataset for Diverse, Explainable Multi-hop Question Answering",
    author = "Yang, Zhilin  and
      Qi, Peng  and
      Zhang, Saizheng  and
      Bengio, Yoshua  and
      Cohen, William  and
      Salakhutdinov, Ruslan  and
      Manning, Christopher D.",
    editor = "Riloff, Ellen  and
      Chiang, David  and
      Hockenmaier, Julia  and
      Tsujii, Jun{'}ichi",
    booktitle = "Proceedings of the 2018 Conference on Empirical Methods in Natural Language Processing",
    month = oct # "-" # nov,
    year = "2018",
    address = "Brussels, Belgium",
    publisher = "Association for Computational Linguistics",
    url = "https://aclanthology.org/D18-1259/",
    doi = "10.18653/v1/D18-1259",
    pages = "2369--2380",
}

@inproceedings{ho20202Wikimultihopqa,
    title = "Constructing A Multi-hop {QA} Dataset for Comprehensive Evaluation of Reasoning Steps",
    author = "Ho, Xanh  and
      Duong Nguyen, Anh-Khoa  and
      Sugawara, Saku  and
      Aizawa, Akiko",
    editor = "Scott, Donia  and
      Bel, Nuria  and
      Zong, Chengqing",
    booktitle = "Proceedings of the 28th International Conference on Computational Linguistics",
    month = dec,
    year = "2020",
    address = "Barcelona, Spain (Online)",
    publisher = "International Committee on Computational Linguistics",
    url = "https://aclanthology.org/2020.coling-main.580/",
    doi = "10.18653/v1/2020.coling-main.580",
    pages = "6609--6625",
}

@article{trivedi2022musique,
    title = "{M}u{S}i{Q}ue: Multihop Questions via Single-hop Question Composition",
    author = "Trivedi, Harsh  and
      Balasubramanian, Niranjan  and
      Khot, Tushar  and
      Sabharwal, Ashish",
    editor = "Roark, Brian  and
      Nenkova, Ani",
    journal = "Transactions of the Association for Computational Linguistics",
    volume = "10",
    year = "2022",
    address = "Cambridge, MA",
    publisher = "MIT Press",
    url = "https://aclanthology.org/2022.tacl-1.31/",
    doi = "10.1162/tacl_a_00475",
    pages = "539--554",
}

@inproceedings{press2023bamboogle,
    title = "Measuring and Narrowing the Compositionality Gap in Language Models",
    author = "Press, Ofir  and
      Zhang, Muru  and
      Min, Sewon  and
      Schmidt, Ludwig  and
      Smith, Noah  and
      Lewis, Mike",
    editor = "Bouamor, Houda  and
      Pino, Juan  and
      Bali, Kalika",
    booktitle = "Findings of the Association for Computational Linguistics: EMNLP 2023",
    month = dec,
    year = "2023",
    address = "Singapore",
    publisher = "Association for Computational Linguistics",
    url = "https://aclanthology.org/2023.findings-emnlp.378/",
    doi = "10.18653/v1/2023.findings-emnlp.378",
    pages = "5687--5711",
}

@article{wikidata,
    author = {Vrande\v{c}i\'{c}, Denny and Kr\"{o}tzsch, Markus},
    title = {Wikidata: a free collaborative knowledgebase},
    year = {2014},
    issue_date = {October 2014},
    publisher = {Association for Computing Machinery},
    address = {New York, NY, USA},
    volume = {57},
    number = {10},
    issn = {0001-0782},
    url = {https://doi.org/10.1145/2629489},
    doi = {10.1145/2629489},
    journal = {Commun. ACM},
    month = sep,
    pages = {78–85},
    numpages = {8}
}

@article{lewis2020goodrag,
  title={Retrieval-augmented generation for knowledge-intensive nlp tasks},
  author={Lewis, Patrick and Perez, Ethan and Piktus, Aleksandra and Petroni, Fabio and Karpukhin, Vladimir and Goyal, Naman and K{\"u}ttler, Heinrich and Lewis, Mike and Yih, Wen-tau and Rockt{\"a}schel, Tim and others},
  journal={Advances in neural information processing systems},
  volume={33},
  pages={9459--9474},
  year={2020}
}

@inproceedings{du2022rag_eae,
    title = "Retrieval-Augmented Generative Question Answering for Event Argument Extraction",
    author = "Du, Xinya  and
      Ji, Heng",
    editor = "Goldberg, Yoav  and
      Kozareva, Zornitsa  and
      Zhang, Yue",
    booktitle = "Proceedings of the 2022 Conference on Empirical Methods in Natural Language Processing",
    month = dec,
    year = "2022",
    address = "Abu Dhabi, United Arab Emirates",
    publisher = "Association for Computational Linguistics",
    url = "https://aclanthology.org/2022.emnlp-main.307/",
    doi = "10.18653/v1/2022.emnlp-main.307",
    pages = "4649--4666",
}

@inproceedings{chen2022rag_murag,
    title = "{M}u{RAG}: Multimodal Retrieval-Augmented Generator for Open Question Answering over Images and Text",
    author = "Chen, Wenhu  and
      Hu, Hexiang  and
      Chen, Xi  and
      Verga, Pat  and
      Cohen, William",
    editor = "Goldberg, Yoav  and
      Kozareva, Zornitsa  and
      Zhang, Yue",
    booktitle = "Proceedings of the 2022 Conference on Empirical Methods in Natural Language Processing",
    month = dec,
    year = "2022",
    address = "Abu Dhabi, United Arab Emirates",
    publisher = "Association for Computational Linguistics",
    url = "https://aclanthology.org/2022.emnlp-main.375/",
    doi = "10.18653/v1/2022.emnlp-main.375",
    pages = "5558--5570",
}

@inproceedings{su2024rag_dragin,
    title = "{DRAGIN}: Dynamic Retrieval Augmented Generation based on the Real-time Information Needs of Large Language Models",
    author = "Su, Weihang  and
      Tang, Yichen  and
      Ai, Qingyao  and
      Wu, Zhijing  and
      Liu, Yiqun",
    editor = "Ku, Lun-Wei  and
      Martins, Andre  and
      Srikumar, Vivek",
    booktitle = "Proceedings of the 62nd Annual Meeting of the Association for Computational Linguistics (Volume 1: Long Papers)",
    month = aug,
    year = "2024",
    address = "Bangkok, Thailand",
    publisher = "Association for Computational Linguistics",
    url = "https://aclanthology.org/2024.acl-long.702/",
    doi = "10.18653/v1/2024.acl-long.702",
    pages = "12991--13013",
}

@inproceedings{li2024rag_longtail,
    title = "On the Role of Long-tail Knowledge in Retrieval Augmented Large Language Models",
    author = "Li, Dongyang  and
      Yan, Junbing  and
      Zhang, Taolin  and
      Wang, Chengyu  and
      He, Xiaofeng  and
      Huang, Longtao  and
      Xue{'}, Hui  and
      Huang, Jun",
    editor = "Ku, Lun-Wei  and
      Martins, Andre  and
      Srikumar, Vivek",
    booktitle = "Proceedings of the 62nd Annual Meeting of the Association for Computational Linguistics (Volume 2: Short Papers)",
    month = aug,
    year = "2024",
    address = "Bangkok, Thailand",
    publisher = "Association for Computational Linguistics",
    url = "https://aclanthology.org/2024.acl-short.12/",
    doi = "10.18653/v1/2024.acl-short.12",
    pages = "120--126",
}

@inproceedings{fang2024rag_enhancing,
    title = "Enhancing Noise Robustness of Retrieval-Augmented Language Models with Adaptive Adversarial Training",
    author = "Fang, Feiteng  and
      Bai, Yuelin  and
      Ni, Shiwen  and
      Yang, Min  and
      Chen, Xiaojun  and
      Xu, Ruifeng",
    editor = "Ku, Lun-Wei  and
      Martins, Andre  and
      Srikumar, Vivek",
    booktitle = "Proceedings of the 62nd Annual Meeting of the Association for Computational Linguistics (Volume 1: Long Papers)",
    month = aug,
    year = "2024",
    address = "Bangkok, Thailand",
    publisher = "Association for Computational Linguistics",
    url = "https://aclanthology.org/2024.acl-long.540/",
    doi = "10.18653/v1/2024.acl-long.540",
    pages = "10028--10039",
}

@inproceedings{zhao2024rag_longrag,
    title = "{L}ong{RAG}: A Dual-Perspective Retrieval-Augmented Generation Paradigm for Long-Context Question Answering",
    author = "Zhao, Qingfei  and
      Wang, Ruobing  and
      Cen, Yukuo  and
      Zha, Daren  and
      Tan, Shicheng  and
      Dong, Yuxiao  and
      Tang, Jie",
    editor = "Al-Onaizan, Yaser  and
      Bansal, Mohit  and
      Chen, Yun-Nung",
    booktitle = "Proceedings of the 2024 Conference on Empirical Methods in Natural Language Processing",
    month = nov,
    year = "2024",
    address = "Miami, Florida, USA",
    publisher = "Association for Computational Linguistics",
    url = "https://aclanthology.org/2024.emnlp-main.1259/",
    doi = "10.18653/v1/2024.emnlp-main.1259",
    pages = "22600--22632",
}

@inproceedings{sankararaman2024rag_provenance,
    title = "Provenance: A Light-weight Fact-checker for Retrieval Augmented {LLM} Generation Output",
    author = "Sankararaman, Hithesh  and
      Yasin, Mohammed Nasheed  and
      Sorensen, Tanner  and
      Bari, Alessandro Di  and
      Stolcke, Andreas",
    editor = "Dernoncourt, Franck  and
      Preo{\c{t}}iuc-Pietro, Daniel  and
      Shimorina, Anastasia",
    booktitle = "Proceedings of the 2024 Conference on Empirical Methods in Natural Language Processing: Industry Track",
    month = nov,
    year = "2024",
    address = "Miami, Florida, US",
    publisher = "Association for Computational Linguistics",
    url = "https://aclanthology.org/2024.emnlp-industry.97/",
    doi = "10.18653/v1/2024.emnlp-industry.97",
    pages = "1305--1313",
}

@inproceedings{mitra2022multi_constraint,
    title = "Constraint-based Multi-hop Question Answering with Knowledge Graph",
    author = "Mitra, Sayantan  and
      Ramnani, Roshni  and
      Sengupta, Shubhashis",
    editor = "Loukina, Anastassia  and
      Gangadharaiah, Rashmi  and
      Min, Bonan",
    booktitle = "Proceedings of the 2022 Conference of the North American Chapter of the Association for Computational Linguistics: Human Language Technologies: Industry Track",
    month = jul,
    year = "2022",
    address = "Hybrid: Seattle, Washington + Online",
    publisher = "Association for Computational Linguistics",
    url = "https://aclanthology.org/2022.naacl-industry.31/",
    doi = "10.18653/v1/2022.naacl-industry.31",
    pages = "280--288",
}

@inproceedings{lee2022multi_gen_multihop,
    title = "Generative Multi-hop Retrieval",
    author = "Lee, Hyunji  and
      Yang, Sohee  and
      Oh, Hanseok  and
      Seo, Minjoon",
    editor = "Goldberg, Yoav  and
      Kozareva, Zornitsa  and
      Zhang, Yue",
    booktitle = "Proceedings of the 2022 Conference on Empirical Methods in Natural Language Processing",
    month = dec,
    year = "2022",
    address = "Abu Dhabi, United Arab Emirates",
    publisher = "Association for Computational Linguistics",
    url = "https://aclanthology.org/2022.emnlp-main.92/",
    doi = "10.18653/v1/2022.emnlp-main.92",
    pages = "1417--1436",
}

@inproceedings{shen2023multi_dense,
    title = "Dense-{ATOMIC}: Towards Densely-connected {ATOMIC} with High Knowledge Coverage and Massive Multi-hop Paths",
    author = "Shen, Xiangqing  and
      Wu, Siwei  and
      Xia, Rui",
    editor = "Rogers, Anna  and
      Boyd-Graber, Jordan  and
      Okazaki, Naoaki",
    booktitle = "Proceedings of the 61st Annual Meeting of the Association for Computational Linguistics (Volume 1: Long Papers)",
    month = jul,
    year = "2023",
    address = "Toronto, Canada",
    publisher = "Association for Computational Linguistics",
    url = "https://aclanthology.org/2023.acl-long.742/",
    doi = "10.18653/v1/2023.acl-long.742",
    pages = "13292--13305",
}
\end{document}